\documentclass{article}
\usepackage{spconf,amsmath,graphicx}

\title{Subject Identification Across large expression variations using 3D facial landmarks}
\name{Sk Rahatul Jannat, Diego Fabiano, Shaun Canavan, and Tempestt Neal}
\address{University of South Florida}

\begin{document}

\maketitle
\begin{abstract}
Landmark localization is an important first step towards geometric based vision research including subject identification. Considering this, we propose to use 3D facial landmarks for the task of subject identification, over a range of expressed emotion. Landmarks are detected, using a Temporal Deformable Shape Model and used to train a Support Vector Machine (SVM), Random Forest (RF), and Long Short-term Memory (LSTM) neural network for subject identification. As we are interested in subject identification with large variations in expression, we conducted experiments on 3 emotion-based databases, namely the BU-4DFE, BP4D, and BP4D+ 3D/4D face databases. We show that our proposed method outperforms current state of the art methods for subject identification on BU-4DFE and BP4D. To the best of our knowledge, this is the first work to investigate subject identification on the BP4D+, resulting in a baseline for the community.
\end{abstract}


\section{Introduction}
Broadly, face recognition can be categorized as holistic, hybrid matching, or feature-based \cite{ZhaoSurvey}. Holistic approaches look at the global similarity of the face such as a 3D morphable model (3DMM) \cite{Blanz3DMM}; hybrid matching make use of either multiple methods \cite{HuangComponent} or multiple modalities \cite{kakadiaris2005multimodal}; feature-based methods look at local features of the face to find similarities \cite{zhong2007robust}. The work proposed in this paper can be categorized as feature-based. Due to its non-intrusive nature and wide applicability in security and defense related fields, face recognition has been actively researched by many groups in recent decades. 

Since some of the earlier methods for face recognition \cite{turk1991face}, \cite{zhao1998discriminant}, to more recent works within the past 10 years \cite{canavan2010evaluation}, \cite{zhang2011sparse} 2D face recognition has been an actively researched field. With the recent advances in deep neural networks, we have seen significant jumps in performance \cite{emambakhsh2016nasal}, \cite{kemelmacher2016megaface}, \cite{parkhi2015deep}, \cite{saragih2011deformable}, \cite{sun2014deep}, \cite{wen2016discriminative}. Liu et al. \cite{liu2017sphereface} proposed the angular softmax that allows convolutional neural networks (CNN) the ability to learn angularly discriminative features. This was proposed to handle the problem where face features are shown to have a smaller intra-class distance compared to inter-class distance. Recently, Tuan et al. \cite{tuan2017regressing} proposed regressing 3D morphable model shape and texture parameters from a 2D image using a CNN. Using this approach, they were able to obtain a sufficient amount of training data for their network showing promising results. Zhu et al. \cite{zhu2015high} proposed a high-fidelity pose and expression normalization method that made use of a 3DMM to generate natural, frontal facing, neutral face images. Using this method, they achieved promising results in both constrained and unconstrained environments (i.e. wild settings). Although performance has been increasing and groups have been actively working on 2D subject identification, there are still some challenges such as pose and lighting. 3D faces can help to minimize these challenges \cite{singh2018techniques}, and in recent years, this research has made significant strides \cite{echeagaray2017method}, \cite{emambakhsh2016nasal}, \cite{soltanpour2017survey}  due to the development of powerful, high-fidelity 3D sensors. 

Echeagaray-Patron et al. \cite{echeagaray2017method} proposed a method for 3D face recognition where conformal mapping is used to map the original face surfaces onto a Riemannian manifold. From the conformal and isometric invariants that they compute, comparisons are then made. This method was shown to have invariance to both expression and pose. Li et al. \cite{li2015towards} proposed the use of SIFT-like matching using three 3D key point descriptors. Each of these descriptors were fused at the feature-level to describe local shapes of detected key points. Lei et al. \cite{lei2014efficient} proposed the Angular Radial Signature for 3D face recognition. This signature is extracted from the semi-rigid regions of the face, followed by mid-level features being extracted from the signature by Kernel Principal Component Analysis. These features were then used to train a support vector machine showing promising results when comparing neutral vs. non-neutral faces. Berretti et al. \cite{berretti20103d} proposed the use of 3D Weighted Walkthroughs with iso-geodesic facial strips for the task of 3D face recognition. They achieved promising results on the FRGC v2.0 \cite{phillips2005overview} and SHREC08 \cite{daoudi2008shrec} 3D facial datasets. Using multistage hybrid alignment algorithms and an annotated face model, Kakadiaris et al. \cite{kakadiaris20063d} used a deformable model framework to show robustness to facial expressions when performing 3D face recognition. 

\begin{figure}[t]
\includegraphics[width=8.5cm, height=4.5cm]{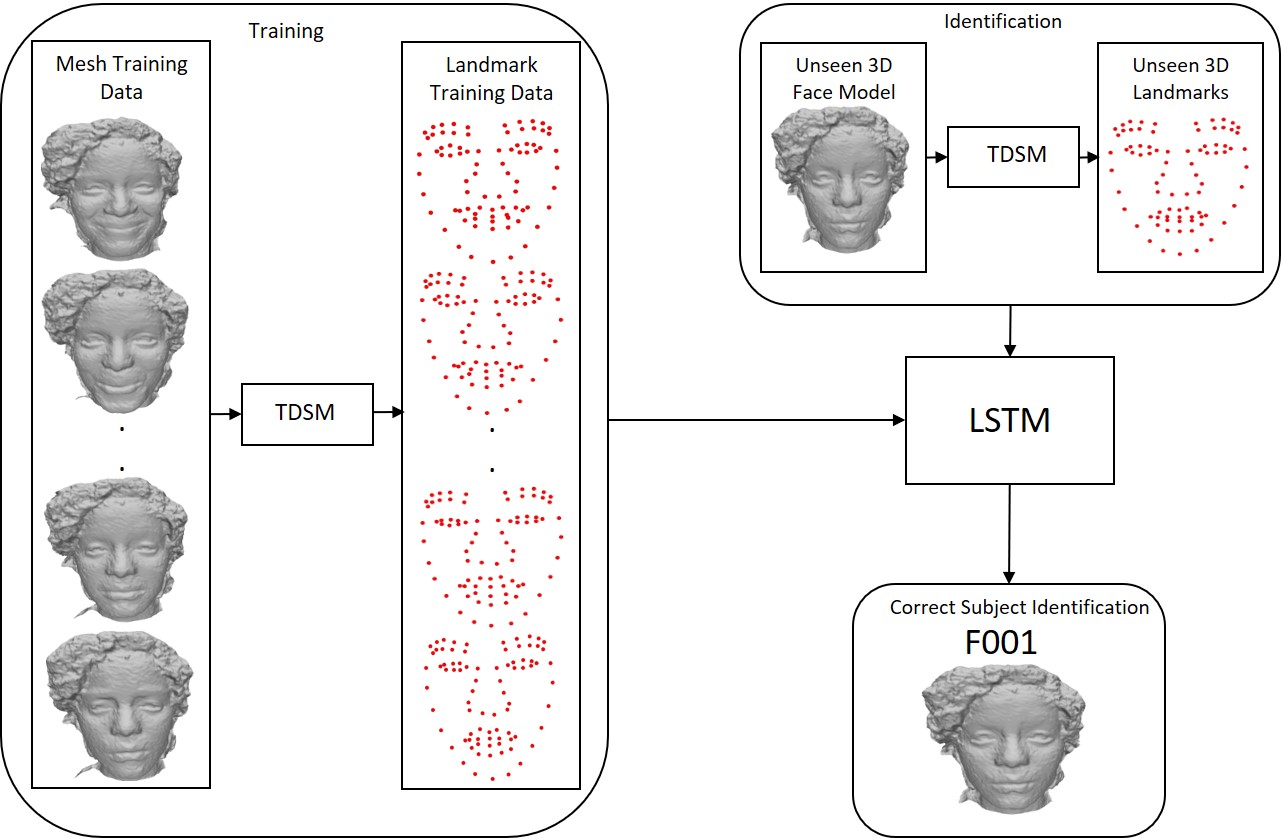}
\caption{Overview of proposed method. Example is showing an unseen 3D mesh model of subject ‘F001’ from BP4D+ \cite{zhang2016multimodal}, who is correctly identified based on training a LSTM \cite{hochreiter1997long} from 3D facial data detected from a TDSM.}
\label{fig:overview}
\end{figure}

Motivated by the above works, we propose to use 3D facial landmarks for subject identification across large variations in expression. We track the facial landmarks using a Temporal Deformable Shape Model (TDSM) \cite{canavan2013fitting}. See Fig. \ref{fig:overview} for an overview of the proposed approach. The rest of the paper is organized as follows. Section \ref{sec:TDSM} gives a brief overview of the TDSM algorithm, Section \ref{sec:subIdResults} details our experimental design and results, and we conclude in Section \ref{sec:conclusion}.

\section{Temporal Deformable Shape Model}
\label{sec:TDSM}
The Temporal Deformable Shape Model (TDSM) models the shape variation of 3D facial data. Given a sequence of data (i.e. 4D), it also models the implicit constraints on shape that are imposed (e.g. small changes in motion and shape). To construct a TDSM, a training set of 3D facial landmarks is required. First, the 3D facial landmarks are aligned using a modified version of Procrustes analysis \cite{de2003adapting}.  Given a training set of size \textit{L} 3D faces, where each face has \textit{N} facial landmarks (aligned with Procrustes analysis), a parameterized model \textit{S} is constructed, $S = F^1_1,...,F^1_N,...,F^m_1,...,F^m_N$. $F^m_i$ is the $i^{th}$ landmarks of the $m^{th}$ 3D face in the training set, where $F^m_i = (x^m_i,y^m_i,z^m_i)$ and $1 \leq m \leq L$. From this model, principal component analysis (PCA), is then applied to learn the modes of variation, \textit{V}, of the training data.

Given the parameterized model, \textit{S}, and the modes of variation, \textit{V}, to detect 3D facial landmarks, an offline weight vector, \textit{w}, is constructed that allows for new face shapes to be constructed (i.e. these face shapes are constructed offline), by a linear combination of landmarks as $S=\Bar{s}+Vw$ where $\Bar{s}$ is the average face shape. These constructed face shapes are constrained to be within the range $-2\sqrt{\lambda_i} \leq w_i \leq 2\sqrt{\lambda_i}$, where $w_i$ is the $i^{th}$ weight in the range, and $\lambda_i$ is the $i^{th}$ eigenvalue from PCA. This constraint is imposed to make sure the new face shape is a 3D face.

To fit (i.e. detect landmarks) to a new input mesh, an offline table of weights (\textit{w}) is constructed with a uniform amount of variance. The Procrustes distance, \textit{D}, is then computed between each face shape (referred to as an instance of the TDSM) and the new input mesh. The smallest distance is considered the best detected landmarks. Note that this is not meant to be an exhaustive overview of a TDSM, therefore we refer the reader to the original work \cite{canavan2013fitting} for more details.
\section{Experimental Design and Results}
\label{sec:subIdResults}
Using a TDSM, we detected 83 facial landmarks on 3 publicly available 3D emotion-based face databases: BU4DFE \cite{yin126high}, BP4D \cite{zhang2014bp4d}, and BP4D+ \cite{zhang2016multimodal}. From these facial landmarks, we then conducted subject identification experiments, where the landmarks are used as training data for 3 machine learning classifiers. Using these 83 facial landmarks we have also reduced the dimensionality of the 3D faces from over 30,000 3D vertices, while still retaining important features for subject identification. This allows us to reduce storage requirements, as well as processing time of the 3D face, which can be limitations of 3D face recognition \cite{bowyer2006survey}, \cite{kakadiaris2007three}. An overview of the databases and the experimental design is detailed in the following subsections.

\subsection{3D face databases}
One of the main goals of this work is to show subject identification across large variations in expression. Considering this, we needed to evaluate large and varied 3D emotion-based face databases. To facilitate this, we chose 3 state-of-the-art 3D emotion-based face databases, and investigated a total of 282 subjects across the 3 datasets.

\textbf{BU-4DFE \cite{yin126high}:} Consists of 101 subjects displaying 6 prototypic facial expressions plus neutral. The dataset has 58 females and 43 males, including a variety of racial ancestries. The age range of the BU-4DFE is 18-45 years of age. 

\textbf{BP4D \cite{zhang2014bp4d}:} Consists of 41 subjects displaying 8 expressions plus neutral. It consists of 23 females and 18 males; 11 Asian, 4 Hispanic, 6 African-American, and 20 Euro-American ethnicities are represented. The age range of the BP4D is 18-29 years of age. This database was developed to explore spatiotemporal features in facial expressions. Due to its large variation in expression, it is a natural fit for our subject identification study. 

\textbf{BP4D+ \cite{zhang2016multimodal}:} Consists of 140 subjects (82 females and 58 males) ages 18-66. This data corpus consists of ethnic and racial ancestries that include African American, Caucasian, and Asian each with highly varied emotions. These emotions are elicited through tasks designed to elicit dynamic emotions in the subjects such as disgust, sadness, pain, and surprise resulting in a challenging dataset. Like the BP4D database, this dataset was also designed to study emotion classification. Its diversity and number of subjects, as well as large variations in expressions, make it a natural fit for our study.

\subsection{Experimental design}
To conduct our experiments, we detected 83 facial landmarks on the 3D data using a TDSM. Given 3D facial landmarks, we then translated them so that the centroid of the face is located at the origin in 3D space to align the data. The translated 3D facial features were then used for subject identification. Each of the 3D facial landmarks ($x$, $y$, $z$ coordinates) are inserted into a new feature vector. For all 83 landmarks, this gives us a feature vector of size $83 \times 3 = 249$. This feature vector is used to train classifiers for subject identification. To ensure our results were not classifier specific, we trained a support vector machine (SVM) \cite{vapnik1998support}, random forest (RF) \cite{breiman2001random}, and Long short-term memory (LSTM) neural network \cite{hochreiter1997long}. Our network consists of one short-term memory layer with a look back of two faces (estimated landmarks), followed by 0.5 dropout, and a fully connected layer for classification. The softmax activation function was used, along with the RMSprop \cite{tieleman2012lecture} optimizer with a learning rate of 0.0001.

For each classifier, each subject’s identity was used as the class (each 3D face is labeled with a subject id). Accurate results on an SVM, RF, and LSTM show the robustness of the 3D facial landmarks to multiple machine learning classifiers. We conducted one-to-many subjection identification, where all subjects were in both the training and testing sets. These sets were split based on time (i.e. different sections of the  sequences available in the datasets) so consecutive (i.e., similar) frames did not appear in both sets.

\subsection{Subject identification results}
We achieved an average  subject identification accuracy of 99.9\%, on random forest and support vector machine, and 99.93\% for a long short-term network, across all databases. As can be seen in Table \ref{table:subID}, an SVM, RF, and LSTM can accurately identify subjects from the BU4DFE, BP4D, and BP4D+ datasets achieving a max accuracy of 100\% on BU4DFE, and a minimum accuracy of 99.8\% on BP4D+. All three of the tested classifiers achieved consistent results across all three datasets, showing these results are not classifier dependent. As each of the datasets contain large variations in expression, these results show the detected 3D landmarks have robustness to expression changes for the task of subject identification. 

\begin{table}[t]
    \caption{Subject identification accuracies for the 3 tested datasets and classifiers.}
    \label{table:subID}
    \centering 
    \begin{tabular}{|c|c|c|c|}
    \hline
     & BU4DFE & BP4D & BP4D+ \\
     \hline
     SVM & 99.9\% & 99.9\% & 99.9\% \\
     \hline
     RF & 100\% & 99.9\% & 99.8\% \\
     \hline
     LSTM & 100\% & 99.9\% & 99.9\% \\
     \hline
    \end{tabular}
\end{table}

\subsection{Subject identification with occluded faces}
\begin{figure}[t]
\includegraphics[width=8.5cm, height=4.5cm]{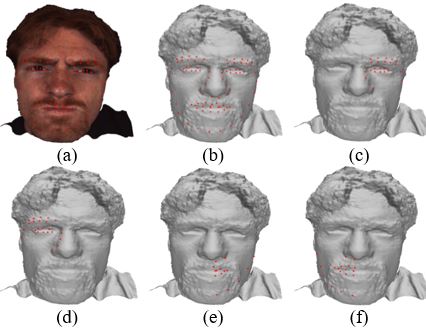}
\caption{Detected landmarks (BP4D \cite{zhang2014bp4d}) used for subject ID (original 3D mesh shown only for display purposes). (a) 83 landmarks with texture (note: texture is shown for display purposes only showing robustness to facial hair); (b) 83 landmarks; (c) top left quadrant; (d) top right quadrant; (e) lower left quadrant; and (f) lower right quadrant.}
\label{fig:occlusion}
\end{figure}

Along with subject identification using all 83 landmarks, we also tested on a smaller number of facial landmarks to simulate occluded faces. For these experiments, we split the 3D facial landmarks (i.e. face) into 4 quadrants (Fig. \ref{fig:occlusion}) and detected a smaller number of landmarks (top right: 23; top left:23; lower right: 20; lower left: 17) using a TDSM. We then ran the same experiments for each quadrant. As shown in Section \ref{sec:subIdResults}, the results are not classifier specific, as the random forest, SVM, and LSTM network have similar results. Due to this we only used a random forest and support vector machine for these experiments.

When testing on simulated occluded faces on BU4DFE, both the random forest and SVM achieved 99.9\% accuracy in all four quadrants, showing robustness to occlusion. Testing on BP4D, the random forest achieved an average accuracy of 99.7\% across the four quadrants, and SVM achieved an average accuracy of 93.2\% across the four quadrants. On BP4D+, random forest and SVM achieved an average accuracy of 99.4\% and 97.5\%, respectively across the four quadrants. These results detail the expressive power of the detected 3D facial landmarks to reliably identify subjects under extreme conditions (e.g. large variations in expression and occlusion). See Table \ref{table:occ} for individual quadrant accuracies for BP4D and BP4D+ (BU4DFE not shown as all quadrants had same accuracy of 99.9\% for both classifiers).
%

\begin{table}[t]
    \caption{Subject identification accuracies(percentage) for faces with simulated occlusion. Key: TR: Top Right; TL: Top Left; LR: Lower Right; LL: Lower Left.}
    \label{table:occ}
    \centering 
    \setlength\tabcolsep{4pt}
    \begin{tabular}{|c|c|c|c|c|c|c|c|c|}
     \hline
     & \multicolumn{4}{|c|}{BP4D} & \multicolumn{4}{|c|}{BP4D+} \\
     \hline
     & TR & TL & LR & LL & TR & TL & LR & LL \\
     \hline
     RF & \bfseries 99.7& \bfseries 99.7 & \bfseries 99.7 & \bfseries 99.7 & \bfseries 99.3 & \bfseries 99.3 & \bfseries 99.6 & \bfseries 99.5 \\
     \hline 
     SVM & 95.1 & 96.8 & 93.4 & 87.5 & 98.8 & 99.1 & 97.5 & 94.8 \\
     \hline
    \end{tabular}
\end{table}
\subsection{Comparisons to state of the art}
We compared our proposed method to the current state of the art on BU-4DFE \cite{yin126high} and BP4D \cite{zhang2014bp4d} (see Table \ref{table:SOA} for both). To the best of our knowledge this is the first study to perform subject identification on BP4D+ \cite{zhang2016multimodal}; therefore, we did not have any works to compare against resulting in a baseline for the community. In these comparisons, it is important to note that Canavan et al \cite{canavan2015landmark} used 1800 and 2400 frames from BU-4DFE and BP4D, respectively, for their experiments. We used all data in both datasets (60402 and 367474 respectively). The work from Sun et al. \cite{sun20083d} also requires both spatial and temporal information to achieve their results of 98.61\%, and while our approach can incorporate temporal information (e.g. LSTM), it can also identify a subject based on one frame of data, which is useful when temporal information is not available.

\begin{table}[t]
    \caption{State-of-the-art comparisons.}
    \label{table:SOA}
    \centering 
    \begin{tabular}{|c|c|c|}
    \hline
     Method & BU4DFE & BP4D \\
     \hline
     Proposed Method (RF) & \textbf{100\%} & \textbf{99.9\%}  \\
     \hline
     Proposed Method (LSTM) & \textbf{100\%} & \textbf{99.9\%}  \\
     \hline
     Proposed Method (SVM) & 99.9\% & \textbf{99.9\%}  \\
     \hline
     Sun et al. \cite{sun20083d} & 98.61\% & N/A \\
     \hline
     Fernandes et al. \cite{lawrence20143d} & 96.71\% & N/A \\
     \hline
     Canavan et al. \cite{canavan2015landmark} & 92.7\% & 93.4\% \\
     \hline
    \end{tabular}
\end{table}
\section{Conclusion}
\label{sec:conclusion}
We have shown 3D facial landmarks can be used for subject identification across large variations in expression. We validated our approach on  three 3D emotion-based face databases (BU4DFE \cite{yin126high}, BP4D \cite{zhang2014bp4d}, and BP4D+ \cite{zhang2016multimodal}), using a random forest, support vector machine, and long short-term neural network. The proposed method outperforms current state of the art on 2 publicly available 3D face databases achieving a max identification accuracy of 100\% on BU-4DFE and 99.9\% on BP4D. To the best of our knowledge, this is the first work to report subject identification results on the BP4D+. We have also shown the detected landmarks can be used for subject identification in the presence of facial occlusion (simulated). We will further investigate this robustness to expression and occlusion in future work, by investigating other state-of-the-art 3D face emotion datasets such as 4DFab \cite{cheng20184dfab}, which was also designed with biometrics studies in mind, as well as large variations in expression.

We are also interested in emotion-invariant multimodal subject identification. In this paper, we have shown that 3D landmarks are invariant to large expression changes for the task of subject identification. Since facial expressions are often physiological responses to emotion, emotion-invariant identification can have a broad range of applications such as medicine and healthcare (e.g., identifying individuals despite expressions of pain). Multimodal approaches are generally more accurate due to the fusion of heterogeneous data, each contributing identifying information. Considering this, we hypothesize a multimodal approach will significantly advance research on emotion-invariant subject identification while yielding new insight on the impact of emotion on novel modalities such as smartphone sensor data (e.g., accelerometer and touch measurements) and other unconstrained and transparently acquired data. Such approaches will be valuable for continuous subject identification.
\bibliographystyle{ieee}
\bibliography{references}

\end{document}